\documentclass{article}
\usepackage{spconf,amsmath,graphicx}
\usepackage{amssymb}
\usepackage{amsbsy}
\usepackage{amsthm}
\usepackage{tabularx}
\usepackage[dvipsnames]{xcolor}
\usepackage{enumitem}
\setlist{nosep, leftmargin=14pt}
\usepackage{multirow}
\usepackage{mwe} % to get dummy images

\title{CO-SEG: AN IMAGE SEGMENTATION FRAMEWORK AGAINST LABEL CORRUPTION}
\name{\small Ziyi Huang$^1$, Haofeng Zhang$^2$,
\small Andrew Laine$^3$, Elsa Angelini$^{3,4}$,  Christine Hendon$^1$, Yu Gan$^5$
}

\address{\small $^1$ Department of Electrical Engineering, Columbia University, New York, NY, USA\\ 
\small $^2$ Department of Industrial Engineering and Operations Research, Columbia University, New York, NY, USA\\
  \small $^3$ Department of Biomedical Engineering, Columbia University, New York, NY, USA\\
  \small $^4$ NIHR Imperial Biomedical Research Centre, ITMAT Data Science Group, Imperial College London, UK\\
\small $^5$ Department of Electrical and Computer Engineering, The University of Alabama, Tuscaloosa, AL, USA\\
	}
\begin{document}
%\ninept
%
\maketitle
\begin{abstract}
Supervised deep learning 
%is poised to play an increasing role in the field of biomedical image analysis. However, its 
performance is heavily tied to the availability of high-quality labels for  training. Neural networks can gradually overfit corrupted labels if  directly trained on noisy datasets, leading to severe performance degradation at test time. 
In this paper, we propose a novel deep learning framework, namely Co-Seg, to collaboratively train segmentation networks on datasets which include low-quality noisy labels. Our approach first trains two networks simultaneously to sift through all samples and obtain a subset with reliable labels. Then, 
%critical for training on small datasets, 
an efficient yet easily-implemented label correction strategy is applied to enrich the reliable subset. Finally, using the updated dataset, we retrain the segmentation network to finalize its parameters. %get the final prediction results. 
Experiments in two noisy labels scenarios demonstrate that our proposed model can achieve results comparable to those obtained from supervised learning trained on the noise-free labels. In addition, our framework can be easily implemented in any segmentation algorithm to increase its robustness to noisy labels.
\end{abstract}
\section{Introduction}
\label{sec:intro}

% Computer-assisted intervention for disease treatment has been long standing challenges. Fundamental in biomedical image analysis, biomedical image segmentation outputs pixel-wise annotations on the target tissue types, providing critical information for diagnosis and treatment. Recently, many algorithms have been designed for biomedical segmentation on multiple imaging modalities \cite{??}. 
Recent years have witnessed an upsurge of interests in biomedical segmentation. Based on fully convolutional networks, U-Net \cite{RFB15a} has been emerging as a classic model which concatenates multi-scale features from the downsampling layers and the upsampling layers. By stacking two U-Net architectures on top of each other, DoubleU-net \cite{jha2020doubleu} is an improved version of U-Net aiming to achieve higher performance on specific tasks. CE-Net \cite{gu2019net} modified U-Net structure by adopting pretrained ResNet blocks in the feature encoding step to capture high-level spatial information. However, these fully supervised learning algorithms are vulnerable to label noise and their performance may be hugely degenerated by noisy labels. Therefore, under noisy labels, it is important to identify and selectively learn from a clean and reliable subset of samples which mainly include data with clean labels, rather than learning from the whole sample set.  

How to improve deep learning performance under noisy labels conditions has caught great attention \cite{frenay2013classification,ma2018dimensionality,goldberger2016training, patrini2017making,han2018co,zhu2019pick}. One direction is to estimate the mathematical model of a noise distribution \cite{goldberger2016training,patrini2017making}. In \cite{patrini2017making}, two procedures were proposed for loss correction and noise transition matrix estimation. 
Another direction is to directly train on clean samples \cite{han2018co,zhu2019pick}. Co-teaching \cite{han2018co} trains two networks simultaneously to pick clean samples for each one. However, most current approaches focus on classification tasks, which cannot be applied to the segmentation where labels are spatially arranged in a dense manner. Finally, sample-based reweighting methods \cite{han2018co, zhu2019pick,mirikharaji2019learning} just ignore or assign small weights on noisy samples, which can lead to severe overfitting, especially for small datasets.

\label{sec:meth}
\begin{figure*}
\begin{center}
\includegraphics[width=0.9\linewidth]{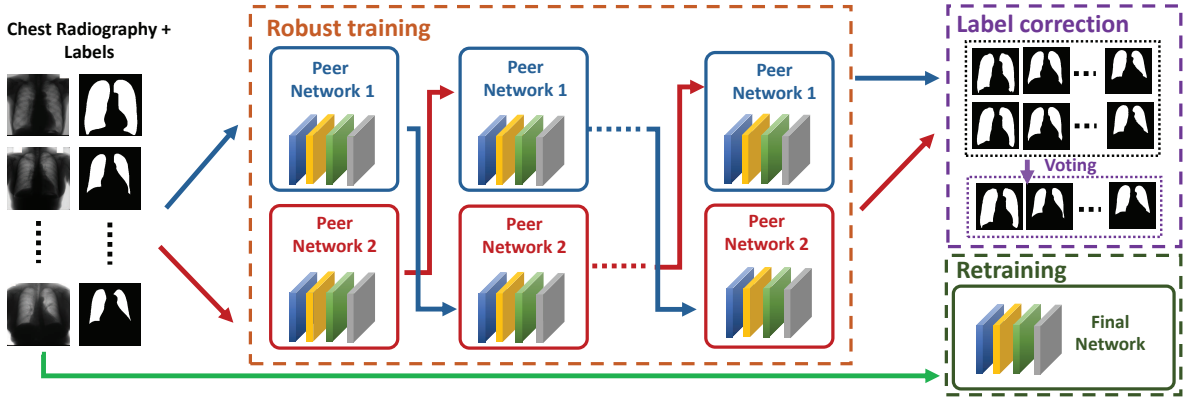}
\end{center}
\vspace{-2em}
\caption{Algorithm flow of Co-Seg framework.} % \textcolor{red}{NEED TO CHANGE TEXT ON FIGURE CT->RADIOGRAPHS }} %Training \textcolor{red}{validation?} Dice (DIC) values
%\vspace{-11pt}
\label{flow}
\vspace{-1em}
\end{figure*}
In this paper, we propose a novel deep learning framework for image segmentation, namely Co-Segmentation (Co-Seg), to handle noisy labels. Our framework integrates the idea of selective training and label correction. In particular, we propose a robust training network to collaboratively learn and select samples with reliable labels. Then a label correction scheme is proposed to enrich the reliable dataset and we retrain a new network on the updated dataset. 
%With an updated high-quality dataset, we are able to train the segmentation network to achieve a higher performance. 
%  Our model consists of two phases, the robust training module and the label correction and retraining module. Our robust learning module uses two networks to learn the representative features by training under clean samples picked by each other. Then, we employ a novel voting strategy to correct the misclassified labels on the detected noisy samples. Finally, a single segmentation network is retrained using the updated labels. We conduct experiments on lung segmentation dataset with two noise types, including a real-world noise scenario generated by an inexperienced human. 
Experimental results using Co-Seg on noisy labels show performance comparable to supervised learning on noise-free labels.
%, demonstrating the learning ability of our model . 
In summary, this paper has the following contributions:\\
(1) We develop an easily-implemented yet effective framework for image segmentation tasks  with noisy labels. It can be easily applied to any deep learning segmentation model to increase learning ability under noisy labels.\\
(2) We demonstrate that, in multiple noise settings, our model achieves comparable results to supervised training on  noise-free datasets. 
\section{Methodology}
Our proposed framework consists of 3 modules: (1) robust training module under noisy labels, (2) label correction module, and (3) retraining module, as shown in Fig. \ref{flow}. The robust learning module trains two segmentation networks simultaneously and selects clean samples for their peer networks. Then, the label correction module employs a voting strategy to correct unreliable labels from the noisy samples measured by corruption scores. Lastly, a single segmentation network is trained on the
updated dataset to finalize network parameters for future segmentation tasks. %provide final prediction results.

\subsection{Robust training against noisy labels}
Robust training against noisy labels is very challenging due to memorization effects of deep learning models. Directly training under noisy labels, the networks can gradually overfit noisy samples. Inspired by \cite{han2018co}, our robust training module collaboratively trains two networks (peer network 1 \& 2) for clean sample selection. Each network picks up a small proportion of high-quality samples in every mini-batch based on sorted corruption scores. Then, such high-quality samples will be fed to its peer network for back propagation. Given a sample, the corruption score $S_{c}$ is:
\vspace{-0.5em}
\begin{equation}\label{CE}
    S_{c}=-\sum_{x\in \Omega}\sum_{l=1}^{L} g_l(x)\log(p_l(x))
\vspace{-0.5em}
\end{equation}
where $L$ is the number of classes, $p_l(x)$ is the estimated probability of class $l$ at pixel position $x\in \Omega$ with $\Omega$ the image domain 
%$\Omega \subset \mathbb{Z}^2$, 
and $g_l(x)$ is the label of the ground truth. The corruption score measures the reliability of the sample. Samples with smaller corruption scores are more likely to be clean. The proportion of samples selected from the corruption score ranking is controlled by $\alpha$, which is related to the noise level of the dataset. 

%The networks back-propagate based on the following loss function respectively. 
The loss function of each segmentation network is a combination of a cross entropy loss $L_{CE}$, a Dice loss $L_{Dice}$, and a $L_2$-regularization term on the parameters $W_f$ of the network:
\begin{align}
    &L_{total} = L_{CE} + \lambda_1 L_{Dice} + \lambda_2 ||W_f||_2^2 \label{all_loss}\\
    &L_{Dice} = 1- \frac{1}{k}\sum_{l=1}^{k}\frac{2\sum\limits_{x \in \Omega}(p_l(x)g_l(x))}{\sum\limits_{x \in \Omega}(p_l(x))^2  + \sum\limits_{x \in \Omega}(g_l(x))^2} %\nonumber  
\end{align}
where $\lambda_1$, $\lambda_2$ are tuned parameters. Here, the Dice loss $L_{Dice}$ is used to capture spatial and structural coherence in the segmentation tasks. %, we employs as part of our loss function. 
\subsection{Label correction}

We propose to correct noisy labels in biomedical segmentation, rather than ignore or downweight them for two reasons. First, maintaining data size is essential. Training on small-size samples may easily lead to severe overfitting. This is particularly important in biomedical applications where reliable labels at expert level are limited. Second, some noisy samples contain pixels with accurate spatial annotations, which could benefit segmentation. %On the other hand, noisy labels still contain pixels with accurate annotations. 
This differs from an image-level classification task where clean and noisy labels are not correlated.

%First, some noisy samples contain pixels with accurate spatial annotations, which could benefit segmentation. This differs from an image-level classification task where clean and noisy labels are not correlated. Second, maintaining data size is essential. Training on small-size samples may easily lead to severe overfitting. This is particularly important in biomedical applications where expert-level reliable labels are limited.

%Therefore, instead of simply ignoring noisy samples, we employ a novel label correction strategy to correct for the misclassified labels in the noisy samples.

In the label correction module, labels in the training set are corrected based on the voting results from the two collaboratively trained networks obtained from the robust training module. First, we differentiate the noisy samples and the clean samples according to their corruption scores ranking. %\textbf{ from the training set \ merged in a single training set.} 
%First, according to the corruption scores ranking, we differentiate the noisy samples (first $\alpha+5\%$) and the clean samples.
For each pixel in the noisy samples, we correct their labels if the prediction results from the two networks are the same but different from the input labels. Our updated dataset consists of clean samples with original labels and noisy samples with corrected labels. This process enriches the clean dataset with noise-corrected samples. 
%\textcolor{red}{EXPLAIN HOW YOU MEASURE AGREEMENT OF THE 2 SEG NETWORK OUTPUTS AND DIVERGENCE FROM THE GROUND TRUTH.}

%Benefiting from the robust training module and the label correction module, now we have a more reliable dataset with less noisy labels. 
\subsection{Retraining}
Based on the updated dataset, we retrain a final segmentation network, which shares the same network structure as one of the peer networks. Similarly, the retraining uses loss function defined in Eq. \ref{all_loss}. Then, this network will be used for future prediction.%the final segmentation prediction is obtained from this network. %\textcolor{red}{NOT CLEAR IF updated dataset  IS ONLY CLEAN DATASETS OR CORRECTED NOISY ONES FROM THE INDIVIDUAL SEG NETWORKS}

%We develop a novel voting mechanism for label correction. After training two neural networks in step 1, we compare the prediction results produced by them. We detect noisy samples according to the following criterion: If the two networks reach an agreement for label prediction, which is yet different from the original label. We update the original labels of noisy samples to the agreement labels.

\section{EXPERIMENTS}
We conduct evaluation on segmentation performance in two noisy settings, including both real-world labeling noise from an inexperienced annotator and synthetic noise.

\begin{figure}[t]
\begin{center}
\includegraphics[width=1.0\linewidth]{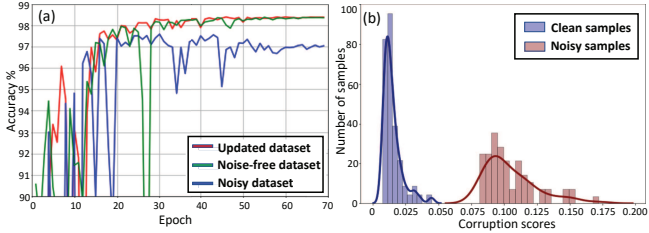}
\end{center}
\vspace{-2em}
\caption{Benchmark U-Net training behavior with 0.5 noise level and Co-Seg corruption scores. (a) U-Net accuracy over training epochs under Type I noise.  (b) Corruption scores under Type II noise: values and probability density functions from the whole training set.}
\vspace{-1em}

% U-Net trained with updated labels achieves higher accuracy and prevents overfitting.

%The corruption scores of the clean samples are very small and clustered in a small range. The probability density functions between clean and corrupted samples have (almost) disjoint supports, indicating that corruption scores effectively separate noisy and clean samples. } % \textcolor{red}{NEED TO CHANGE TEXT ON FIGURE CT->RADIOGRAPHS }} %Training \textcolor{red}{validation?} Dice (DIC) values
%\vspace{-11pt}
\label{loss}
\end{figure}

\textbf{Datatsets.} Our evaluation is based on Chest X-ray from the Japanese Society of Radiological Technology (JSRT) dataset \cite{shiraishi2000development}. This dataset consists of 247 posterior-anterior (PA) chest radiographs. Ground-truth lung masks were obtained from the annotation of Radiographs (SCR) database \cite{van2006segmentation} at expert level. Following previous work in \cite{he2019non,hwang2017accurate}, we resize all images into $256 \times 256$ pixels and split the training and testing sets by ID number: the training set contains 124 images with odd ID number and the testing set contains 123 images with even ID.

\textbf{Label corruption scenarios.} We conduct experiments using two scenarios of label corruption. (1) Synthetic boundary (Type I) noise. Manual segmentation variability usually occurs around tissue boundaries due to spatial uncertainty of contrast transition between different tissue types. Following \cite{zhu2019pick}, we generate boundary noise by randomly eroding or dilating tissue boundaries by $ n_i$ pixels with $1 \le n_i \le 8$ in each direction; (2) Inexperienced annotation (Type II) noise. Human annotators tend to have a systematic bias as over and under segmented tissue boundaries. To mimic this real-world noise scenario, an inexperienced annotator who was blind to the algorithm and ground truth, manually labeled the data and generated biased labels.

%: (1) the boundary noise and (2) the inexperience human noise \textbf{better name for (1) and (2)}. The noise in biomedical image segmentation always occurs around the tissue boundary. So, following the previous research \cite{zhu2019pick}, we also randomly selected 0\%, 25\%, 50\% and 75\% samples from the training set and further randomly eroded or dilated them with $1 \le n_i \le 8$ and $5 \le n_i\le 13$ pixels. 2) \textbf{inexperienced human labeling errors}. A non-clinical-expert rater labeled the images to mimic this real-world noise scenario. No noise level control.
%In addition, we use the percentage of noisy labels in the training set to evaluate the effectiveness of the label correction module. Two scenarios of noise corruption are used to evaluate the performance of our framework

\textbf{Experimental setup.}
%Our proposed framework can be easily implemented on multiple deep learning networks for noise immunity. 
As a demonstration of the framework, we choose the classic U-Net \cite{RFB15a} as the network architecture to evaluate noise robustness performance and we adopt the same U-Net architecture and hyper-parameters for all segmentation networks in the robust training module (peer network 1 and 2 in Fig. \ref{flow}) and retraining module (final network). We also use the performance of a single U-Net segmentation network with the same architecture on noise-free training set as a baseline for comparison. The segmentation performance is evaluated by both accuracy (ACC) and Dice coefficient (DIC) in comparison with ground truth. The corrupted training sets are generated by replacing a proportion of clean samples with noisy samples. %controlling for the noise level ($NoL$) defined as the probability of samples in the training set to be corrupted.
The noise level ($NoL$) is defined as the proportion of noisy samples in the training set. % {\color{red} For Type I error, we synthetically set the $NoL$ of pixels in diluted region as corrupted. For Type II error, we randomly pick up $NoL$ of scans and replaced their labels by the labels obtained from inexperienced annotator.} 
%For noisy sample generation, we use noise level to indicate the probability of samples to be corrupted.
Since the noise level in real-world datasets is around $8\% \sim 38\%$ \cite{song2020learning}, we conduct experiments under 0.1 to 0.5 noise levels for each noise type. Following previous research \cite{han2018co,zhu2019pick}, we assume $NoL$ is a known parameter and $\alpha = 1 - NoL.$  %We report performance on networks trained with noisy labels corrupted with one of three methods, but always compared to clean labels at test time.

% \begin{table}[htb]
% \small
% \vspace{-1em}
% \centering
% \caption{Evaluation metrics on lung segmentation with different noise types (Type I and II explained in the text) and noise levels (NoL) expressed as proportion of noisy labels. ACC=accuracy, DIC=Dice coefficient.  %\textcolor{red}{WHY ARE SOME VALUES  MISSING FOR TYPE II?}. 
% Values in bold indicate best Co-Seg performance and benchmark performance.}
% \begin{tabular}{|c|c|c|c|c|c|c|c|}
% \hline
% \multicolumn{2}{|c|}{ } & \multicolumn{5}{c|}{Co-Seg}\\ \hline

% %\cline{2-9} \cline{10-12}
% \multicolumn{2}{|c|}{Noise Type}  & 0.1 & 0.2 & 0.3 & 0.4 & 0.5 \\ \hline
% Type & ACC &0.9780  &0.9751   & 0.9778 &0.9783  & \textbf{0.9801}\\
% I & DIC  &\textbf{0.9753} &0.9730  & 0.9751 &0.9740  & \textbf{0.9759}\\ \hline
% Type &ACC &  & 0.9800  & 0.9803  & \textbf{0.9798} &  0.9779 \\
% II &DIC  & & 0.9736  & 0.9745   &\textbf{0.9748} & 0.9728\\
% \hline
% \multicolumn{2}{|c|}{Noise-free U-Net} & \multicolumn{5}{c|}{ACC: \textbf{0.9806}; DIC: \textbf{0.9761}}\\
% \hline
% \end{tabular}
% %Percentage of noisy pixels before and after iterations in the training set (without clean samples), superpixel size = 1000.
% \label{correct}
% \end{table}

%\subsection{Experiments under Boundary Noise Corruption}

\textbf{Results.} Table \ref{correct} reports evaluation metrics from the JSRT datasets on lung segmentation with different noise types and noise levels, together with the baseline experiment on noise-free dataset using U-Net. %, which is considered as the benchmark upper bound. 
%\textcolor{red}{EXPLAIN IF U-Net IS ALSO USED IN CO-SEG? SAME ARCHITECTURE/DEPTH OR OPTIMIZED FOR CO-SEG TO MATCH PERFORMANCE? WHAT ABOUT CO-SEG ON NOISE-FREE SAMPLES? DOES IT RECOGNIZE THAT NO SAMPLES NEED TO BE CORRECTED?}. 
The results obtained by our model are comparable with the baseline noise-free U-Net training. Differences are all below $0.6\%$ in both DIC and ACC. Small variations among noise levels are likely caused by model stochasticity. Those results demonstrate that our Co-Seg model can provide robust results under noise levels up to 0.5 with performance competitive with learning using noise-free labels. 

\begin{table}[t]
\footnotesize
\vspace{-1em}
\centering
\caption{Evaluation metrics on lung segmentation with different noise types (Type I and II explained in the text) and noise levels expressed as the proportion of noisy samples. }%ACC=accuracy, DIC=Dice coefficient.}  %\textcolor{red}{WHY ARE SOME VALUES  MISSING FOR TYPE II?}. 
%Values in bold indicate best Co-Seg performance and benchmark performance.}
\begin{tabular}{cccccccc}
\hline
\multirow{2}{*}{Network} & \multirow{2}{*}{Noise}      & \multirow{2}{*}{Metrics} & \multicolumn{5}{c}{Noise level for Type I and II}       \\ \cline{4-8} 
                         &                             &                          & 0.1   & 0.2   & 0.3   & 0.4   & 0.5   \\ \hline
\multirow{4}{*}{Co-Seg}  & \multirow{2}{*}{Type I}     & ACC                      & 0.978 & 0.975 & 0.978 & 0.978 & 0.980 \\ \cline{3-8} 
                         &                             & DIC                      & 0.975 & 0.973 & 0.975 & 0.974 & 0.976 \\ \cline{2-8} 
                         & \multirow{2}{*}{Type II}    & ACC                      &  0.981     & 0.980 & 0.980 & 0.980 & 0.978 \\ \cline{3-8} 
                         &                             & DIC                      &  0.974     & 0.974 & 0.975 & 0.975 & 0.973 \\ \hline
\multirow{2}{*}{U-Net}   & \multirow{2}{*}{Noise Free} & ACC                      & \multicolumn{5}{c}{0.981}             \\ \cline{3-8} 
                         &                             & DIC                      & \multicolumn{5}{c}{0.976}             \\ \hline
\end{tabular}
%Percentage of noisy pixels before and after iterations in the training set (without clean samples), superpixel size = 1000.
\vspace{-1em}
\label{correct}
\end{table}

%\textbf{Effectiveness of corruption score.} 
We further visualize experimental results to show the effectiveness of the Co-Seg model.
%In addition, compared with the noisy samples, the corruption scores for the clean samples are very small and they are clustered in a small range. Therefore, it can directly separate the noisy sample from the clean samples, showing the effectiveness of our noise sample detection. 
Fig. \ref{loss}(a) compares the training accuracy curves (background and lung segmentation) from the benchmark U-Net trained using the updated dataset, noise-free dataset and the noisy dataset over training epochs. Accuracy with the noisy dataset (blue) decreases after 30 epochs % \textcolor{red}{NEED TO CORRECT FIGURE, ADD COLOR LEGEND AND REVISE COMMENTS HERE}, 
indicating that the network is gradually overfitting the noisy samples. Meanwhile, the accuracy curve with the updated training set (red) is flat and smooth, showing consistency of training quality similar to the curve from noise-free dataset (green). Fig. \ref{loss}(b) shows the corruption scores and the probability density functions from the entire training set with 0.5 noise level under Type II label noise corruption. The blue/red curves are the probability density functions fitted on the clean/noisy samples. The two probability density functions have (almost) disjoint supports, indicating that corruption scores effectively separate noisy and clean samples. 

%In Figure \ref{loss}, we report the loss values for noise samples and clean samples. In Figure \ref{ACDIC}, we report accuracy and Dice coefficient value for testing set over the training epoch. 

\begin{figure}
\begin{center}
\includegraphics[width=1.0\linewidth]{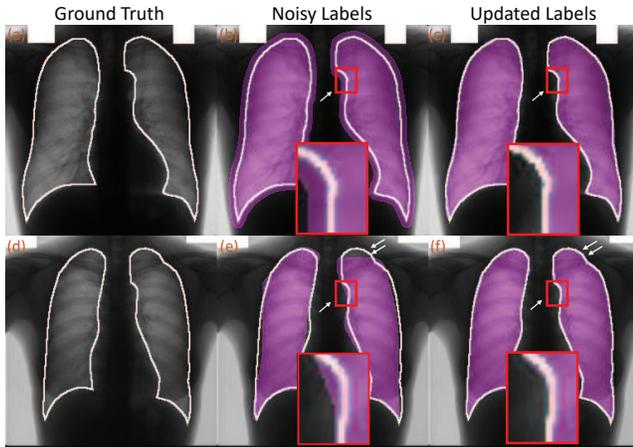}
\end{center}
\vspace{-2em}
\caption{Label correction results with 0.5 noise level for Type I (top row) and Type II (bottom row) noise types. Ground-truth segmentation lung  boundaries are shown in white while the noisy/updated labels are marked in pink.} %The label correction module can accurately amend misclassified boundary pixels. } % \textcolor{red}{NEED TO CHANGE TEXT ON FIGURE CT->RADIOGRAPHS }} %Training \textcolor{red}{validation?} Dice (DIC) values
%\vspace{-11pt}
\vspace{-1em}
\label{EvaluationU-Net}
\end{figure}

Figure \ref{EvaluationU-Net} shows the effect of our label correction module in the two noise scenarios. 
%The updated labels are consistent with ground truth contours. 
In Fig. \ref{EvaluationU-Net} (top row), the Type I noisy labels are well corrected. In Fig. \ref{EvaluationU-Net} (bottom row), the Type II noisy labels are all well corrected at boundary locations marked by a single arrow. In addition, Co-Seg also fills in the large region of missing pixels marked by a double arrow. %Therefore, our model has the ability to learn in real-world noise scenarios.
\section{Conclusions}
\label{sec:Conclusions}
In this paper, we develop a novel collaborative training framework, Co-Seg, to improve segmentation robustness against noisy labels. The robust training module uses two networks to learn the representative features from reliable samples in a noisy dataset. A label correction module employs a voting mechanism to enrich a reliable set prior to final retraining. Experimental results in both synthetic and real-world noise scenarios show that our Co-Seg model is robust to label corruption and achieves comparable results with those trained with noise-free datasets. Importantly, our training scheme is generic and can be easily applied to other deep learning models to increase noise immunity. Future work will focus on extensive validation on more medical segmentation tasks.

\small \section{Compliance with Ethical Standards} 
This research study	 was conducted retrospectively using	 human subject data made available in open access by the Japanese Society of Radiological Technology. Ethical	approval was not required as confirmed by the license	 attached with the open access data.

\section{Acknowledgments}
\label{sec:acknowledgments}\small
The study was funded in part by the National Institute of Health (4DP2HL127776-02, CPH, subaward of UL1TR003096, YG), National Science Foundation (CRII-1948540, YG).

%\begin{figure}
%\begin{center}
%\includegraphics[width=1.0\linewidth]{Picture2.png}
%\end{center}
%\vspace{-2em}
%\caption{Accuracy and Dice coefficient value for testing set along the training epoch when $NL = 0.5, \ n_i \in [5, 13]$. The segmentation accuracy reaches around 96\% and Dice coefficient reaches around 95\%, showing that the model has the ability to get converged to a high value against noisy samples.} % \textcolor{red}{NEED TO CHANGE TEXT ON FIGURE CT->RADIOGRAPHS }} %Training \textcolor{red}{validation?} Dice (DIC) values
%\vspace{-11pt}
%\label{ACDIC}
%\end{figure}
\bibliographystyle{IEEEbib}
\small \bibliography{refs}

%\section{Compliance with Ethical Standards}
%This research study	 was conducted retrospectively using	 human subject data made available in open access by the Japanese Society of Radiological Technology. Ethical	approval was not required as confirmed by the license	 attached with the open access data.

\end{document}